\documentclass{article}

\usepackage[final]{aiplans_2021}

\usepackage[utf8]{inputenc} 
\usepackage[T1]{fontenc}    
\usepackage{hyperref}       
\usepackage{url}            
\usepackage{booktabs}       
\usepackage{amsfonts}       
\usepackage{nicefrac}       
\usepackage{microtype}      
\usepackage{xcolor}         
\usepackage{float}
\usepackage{graphicx}
\usepackage{xcolor}
\usepackage[inline]{enumitem}
\usepackage{amssymb}
\usepackage{pifont}
\newcommand{\cmark}{\ding{51}}%
\newcommand{\xmark}{\ding{55}}%
\usepackage{longtable}
\graphicspath{{images/}{../images/}}
\usepackage{caption} 
\usepackage{listings}
\usepackage{float}
\usepackage{bm}
\usepackage{listings}
\usepackage{wrapfig}
\usepackage{comment}
\usepackage{fancyhdr}
\title{Learning C to x86 Translation: An Experiment in Neural Compilation}

\author{Jordi Armengol-Estapé \& Michael F.P. O'Boyle \\
School of Informatics\\
University of Edinburgh\\
\texttt{jordi.armengol.estape@ed.ac.uk, mob@inf.ed.ac.uk} \\
}

\begin{document}

  \maketitle
  \thispagestyle{fancy}

\begin{abstract}
Deep learning has had a significant impact on many fields. Recently, code-to-code neural models have been used in  code translation, code refinement and decompilation. However, the question of whether these models can automate compilation has yet to be investigated. In this work, we explore neural compilation, building and evaluating Transformer models that learn how to produce x86 assembler from C code.
Although preliminary results are relatively weak, we make our data, models and code publicly available to encourage further research in this area.
\end{abstract}

\section{Introduction}


Machine learning based compilation has been explored 
for over a decade \citep{wang2018machine}. Early work focused on learning profitability
heuristics while more recently, 
deep learning models have been used to build code-to-code models,  for translating or decompiling code. However, to the best of our knowledge, there has  been no prior work on using machine learning
to entirely automate compilation, i.e given a high level source code program generate the equivalent assembler code.
Compilers are large, complex objects \citep{10.5555/977395.977673} and automating their behavior 
represents a significant research challenge.

In this paper, we investigate whether it is possible to learn an end-to-end machine compiler using neural machine translation. In particular, we focus on the translation of small C functions to x86
assembler.
Given that there has been over 50 years research into developing
reliable compiler technology, it may seem unnecessary to expend
further effort in learning a solved problem. However, using a neural translation
approach opens up the possibility of unsupervised compilation from a
language to an ISA (instructon set architecture/machine code). This means we may be
able to automatically generate compilers for new programming languages
and new hardware. 
If true, this enables programming language
researchers and hardware architects to rapidly explore new designs
and will have a transformational impact on both domains. Indeed, there is  already work in unsupervised translation between
programming languages \citep{lachaux2020unsupervised}. However,
before we can begin to consider unsupervised compilation, we first
have to determione whether supervised neural compilation is feasible.

To learn this C $\rightarrow$ x86 translation, we use an existing function-level C corpus, Anghabench \citep{9370322}, to build a parallel C-x86 assembler corpus. Then, we model the compilation task as a sequence-to-sequence task (akin to machine translation) with the Transformer architecture \citep{DBLP:journals/corr/VaswaniSPUJGKP17}. We study the effect of modifying different settings by varying training data size, model size, number of epochs, and other hyperparameters. 
While we can  successfully generate syntactically correct assembler over 80\% of the time and obtain high BLEU scores (c. 90 BLEU in some benchmarks), generating 
semantically correct assembler is more challenging.

The best model can only compile correctly about 33\% of the functions in a benchmark built from an existing program synthesis evaluation set \citep{DBLP:journals/corr/abs-2010-04811}; it specially struggles to compile functions with numerous arguments and arrays.


The article is structured as follows. In Section \ref{sec:related}, we briefly  summarise   related work in NLP and machine learning for code. In Section \ref{sec:method}, we formalize the task of machine compilation and propose how to effectively build neural compilers and fairly evaluate them. Then, in Section \ref{sec:experiments} we establish our experimental framework and report results. Finally, we discuss our approach and conclude, in sections \ref{sec:discuss} and \ref{sec:con}, respectively. 

\section{Related Work}
\label{sec:related}


\paragraph{Natural Language Processing and Machine Translation}  In Natural Language Processing (NLP), the current state-of-the-art typically involves using some variant of the Transformer architecture \citep{DBLP:journals/corr/VaswaniSPUJGKP17} together with some form of subword tokenization \citep{DBLP:journals/corr/SennrichHB15}. In this work, we omit the prolific literature in unsupervised NLP (most famously, BERT \citep{DBLP:journals/corr/abs-1810-04805}) and machine translation because we focus on a supervised setting.


\paragraph{Deep Learning for Code and Symbolic Data} Recent works have proposed to use the encoder-decoder Transformer architecture out of the box for symbolic mathematics \citep{DBLP:journals/corr/abs-1904-01557, DBLP:journals/corr/abs-1912-01412}, or even for automated symbolic proving with decoder-only Transformers \citep{DBLP:journals/corr/abs-2009-03393}. The state-of-art NLP systems for unsupervised pretraining have also been with successfully applied to code, as in CodeBERT \citep{feng2020codebert}. However, other research lines explore the use of alternative modeling strategies for code instead of flat sequences, such as trees to leverage the grammar \citep{DBLP:journals/corr/abs-1802-03691} or other kinds of graphs for data flow analysis \citep{cummins2021programl}.

\paragraph{Machine Learning for Compilers} Many works have proposed the use of machine learning for performance improvement\citep{emani2015celebrating}
or code optimization \citep{9232934}. The field is gaining momentum with recent works such as the CompilerGym \citep{CompilerGym}, a reinforcement learning environment for compilers optimization. However, the common approach is to use machine learning for decision-making e.g. \cite{cavazos2006hybrid}, not to directly generate assembler with a machine learning decoder.

\paragraph{Code Translation and Code-to-Code models} Code-to-code models have been applied in tasks such as \begin{enumerate*}
    \item programming language translation \cite{drissi2018program}, even in unsupervised settings \cite{lachaux2020unsupervised},
    \item code refinement \cite{DBLP:journals/corr/abs-1812-08693}, or
    \item decompilation \cite{DBLP:journals/corr/abs-1905-08325}
\end{enumerate*}.
The latter is  the inverse task of the one we are posing,
and in this specific work it was constrained to a highly restricted subset of C with a maximum of 5 statements.

To the best of our knowledge, no previous work has addressed the task of machine compilation.
One specific additional challenge we note, 
is that 
the target sequences are considerably longer (being assembler) instead of a similar size (as usual in machine translation) or considerably shorter (as usual in summarization or decompilation).


\section{Methods}
\label{sec:method}

We pose machine compilation as a sequence-to-sequence task. Akin to machine translation, machine compilation is the task of translating code into assembler language. More formally, given a dataset $\mathbf{D}$ with $N$ pairs $(\mathbf{x_i},\mathbf{y_i})$, where $\mathbf{x_i}$ is an input program and $\mathbf{y_i}$ is the corresponding assembler code, 
the system is trained with max likelihood estimation: $\ell(\bm{\theta},\mathbf{D}) = \sum_{(\mathbf{x_i},\mathbf{y_i}\in \mathbf{D})} ln  p(\mathbf{y_i}|\mathbf{x_i},\bm{\theta})$. Posing the task as a sequence-to-sequence task conditions both the data generation and the model building.


%

\subsection{Training data}

Regarding the granularity, as a first approach we decided to consider functions, following \citet{lachaux2020unsupervised}. Functions, unlike statements, are standalone units of meaning that can be translated, but at the same they are shorter and easier to test (unit tests) than a whole program (integration tests).  Since  we investigate a supervised setting, we need pairs (C functions, x86 assembler). However, C functions cannot be directly compiled; they typically need additional context (inclusion of headers, type definitions, constant definitions). Thus, even if we have pre-exiting C compilers, generating these data pairs is not trivial.

For this work, we base our dataset on Anghabench \citep{9370322}, a benchmark of around 1 million C functions coupled with the required minimal C code to compile them. Anghabench is built by crawling C code from Github repositories. The authors extracted individual functions, and applied type-inference to reconstruct the missing definitions required to compile them (e.g., declarations of auxiliary functions, type definitions, etc). However, while these reconstructions makes the functions compilable, they are not executable. Apart from not necessarily having a main function and input/output calls, the declared auxiliary functions are not defined. This, among other issues, prevents execution.

It is not practical to directly use this dataset for neural compilation. The inclusion of headers and type definitions while necessary for GCC, adds noise to the machine translation task. We refer to Appendix \ref{app:preprocessing} for the description of our preprocessing pipeline together with the statistics (Appendix \ref{app:data}) of the resulting dataset, which we call Angha-Par (Angha Parallel).  After filtering for length, we kept as many as 500k programs (Angha-Par500k) and a subset (250k) of those for an ablation study (Angha-Par250k).



\subsection{Evaluation}
\label{sec:eval}


\paragraph{BLEU} Machine translation is usually evaluated with BLEU score \citep{10.3115/1073083.1073135}, based on n-gram overlaps between the generated sequence and the ground truth one (in our case, the GCC assembler). This metric does not take into account syntactical or semantic correctness. However, it is easy to compute for all cases.

\paragraph{Syntactic accuracy} We use GCC to check if the assembler generated is syntactically correct, by asking it to generate object code from the  assembler. This metric is more relevant and even easier to compute.

\paragraph{IO accuracy} We evaluate semantic correctness using  \textit{observational equivalence} or \textit{IO accuracy} between the reference GCC assembler and the one output by the model, following recent works on program translation \citep{lachaux2020unsupervised}. That is, we check whether for a given set of inputs, the assembler predicted by the models have the same output as the reference GCC compilation (in other words, we evaluated whether the assembler functions generated by the models pass the available unit tests). While this is no proof that the two programs are formally equivalent, in practice it is a high indicator that it is. This is the most relevant metric and the one we use for selecting the best model and assessing its real performance.   However, Anghabench programs, while compilable, are not executable, since the function dependencies are not included. Thus, we cannot run unit tests on them. For this reason, we take a subset\footnote{Arbitrarily selected based on difficulty of evaluation implementation.} of 64 functions extracted from the program synthesis benchmark collated in \citet{DBLP:journals/corr/abs-2010-04811}. We then add a \texttt{main} function with the required input/output calls to execute them with randomly generated input/output pairs (referred as IO examples, from now on). We refer to this benchmark as Synthesis-bench.

\subsection{Model}

Following previous work on machine translation and deep learning for symbolic mathematics and source code modeling, we use a Transformer model (encoder-decoder) in different settings. We implement all models with Fairseq \citep{ott2019fairseq}, a PyTorch \citep{NEURIPS2019_9015} sequence modeling library. As usual in sequence-to-sequence models, we train with teacher forcing and use a special token to denote the end of the sequence, which is also predicted by the model.

\subsection{Code and Data Availability}

We release\footnote{At \url{https://github.com/jordiae/neural-compilers}} both the code and the data used in this work for the sake of reproducibility.

\section{Experiments}
\label{sec:experiments}



We experiment with 4 Transformer sizes (Trans-Small, Trans-Med-Trans-Big, Trans-Big+). With Trans-Big we further experiment with different vocabulary and data sizes, and number of epochs. See Table \ref{tab:results-tasks} for the parameter count of each of the models, and Appendix \ref{app:experiments} for more details on the models.  We train all models with the same data (except from the ones that have a different vocabulary, which use a different tokenizer, and the one that uses half of data) and the same number of epochs, 5 (except for the model additionally trained for 5 more epochs).  Regarding other hyperparameters, all models are trained with the Adam optimizer \citep{kingma2017adam}. We refer to Fairseq and our source code for additional details. We do not conduct any hyperparameter search, aside from the different configurations reported in Table \ref{tab:results-tasks}. We then evaluate as described in Section \ref{sec:eval}, always with beam search ($k = 5$) and taking the best hypothesis among the top 5. 


\begin{table}
\centering
\scalebox{1.0}{
\begin{tabular}{lr|ccc|cc}
\hline
& & & \textsc{Synthesis}  & & \textsc{Angha-Par} & \\
\hline
\textsc{Model} & \textsc{Params} & \textsc{IO} & \textsc{Syntax} & \textsc{BLEU} & \textsc{Syntax} & \textsc{BLEU}\\
\hline
Trans-Small &    30.9M                            & \phantom{0}0/64                            & \phantom{0}0/64                             & 32.68      &  \textbf{98.50}  & 47.53                \\
Trans-Med & 142.7M & 18/64 & 35/64 & 77.99 & 81.60 & 89.52 \\
Trans-Big  &  193.1M                               & 19/64                           & \textbf{37/64}  & 78.03  & 82.70  & 89.20                        \\
\hspace{1mm} - 50\% data &     193.1M    & 13/64                           & 34/64                            & 76.81      & 88.16 & 83.80                    \\
\hspace{1mm} - 1/2x vocab &    184.7M   & 19/64                           & 36/64                            & 78.07    & 75.60 & 88.63                        \\
\hspace{1mm} + 1/2x vocab  &    209.8M  & 20/64                           & 36/64                            & \textbf{79.48} & 79.30 & 89.21 \\
\hspace{1mm} + 1e2x w.-decay & 193.1M & 18/64                           & 34/64                            & 77.73  & 82.00 & 89.55                         \\
\hspace{1mm} \textbf{+ 1/2x epochs} &   193.1M      & \textbf{21/64} & \textbf{37/64}  & 78.10  & 82.50   & \textbf{90.19}                          \\
Trans-Big+ &  251.9M                                & 19/64                           & 34/64                            & 78.19   & 82.50 & 89.76 \\

\end{tabular}} 
\caption{\label{tab:results-tasks}
Results summary. For each model (namely, the small Transformer variant, the medium-size Transformer, the bigger Transformer variant, the latter plus varying training data size, vocabulary size, with additional weight decay regularization, and with additional training iterations, and an even bigger Transformer variant) we show the total parameter count and report their results in Synthesis-Bench and the Angha-Par test set. Specifically, we report the correct IO examples in Synthesis-Bench, and the syntactic accuracy and BLEU score in Synthesis-Bench and the Angha-Par test. The syntactic accuracy is reported as a fraction for Synthesis-Bench and as a percentage for Angha-Par (due to having a considerably larger number of instances).  In bold, the best results for each metric and dataset, and the best model (Transformer-Big + 1/2 epochs) as per the most relevant variant, correct IO examples.
}
\end{table}



Table \ref{tab:results-tasks} shows the results summary of each model, together with their respective size. The best model, as per the most relevant metric (IO evaluation, that is, observational equivalence) is the Transformer-Big trained for 10 epochs. 

\section{Discussion and Conclusion}
\label{sec:discuss}

\paragraph{Results}
Transformer-Big+1/2x epochs is the best model in terms of the most relevant metric, IO accuracy (observational equivalence). It is the best model in terms of syntactical accuracy in Synthesis-Bench and BLEU score in the Angha-Par test. The smallest model clearly underfits the task of machine compilation, while all reasonably sized models achieve similar enough results (except the model trained with half of the data, which performs considerably worse). 

\paragraph{On the surprisingly high syntax accuracy of the small model}

The smallest model variant, Transformer-Small
obtains
a surprisingly high syntactic accuracy in Angha-Par, as shown in Table \ref{tab:results-tasks},  
 given that 
that its outputs are abnormally long (see Table \ref{tab:lengths}. On inspection
the outputs do not correlate with the inputs. Instead, the model behaves as a nunconditional assembler language model. Furthermore, we observe repeated outputs with different inputs. This phenomenon is reminiscent of the \textit{hallucinations} described in 
other Sequence-to-Sequence models \citep{raunak2021curious} and the \textit{mode collapse} of some generative models \citep{DBLP:journals/corr/abs-1807-04015}.





\paragraph{Error analysis}
Focusing on the outputs of the best model, we observe that \begin{enumerate*}
    \item if a model has one correct IO example for a given function, then it is highly  likely that other IO examples are correct  \item many syntactical errors occur because of a premature end of the hypothesis, \item IO accuracy does not correlate with cyclomatic complexity,\footnote{See Appendix \ref{app:results} for the definition of cyclomatic complexity, a well-known complexity measure.} but with the number of function arguments and pointer variables in the function (as shown in Table \ref{tab:corr}), \item models fail and succeed in the same functions, \item correct model outputs look very similar to the GCC ones although not necessarily identical, \item there are some trivial errors, such as \texttt{true} and \texttt{false} being confused with variable names instead of boolean values
\end{enumerate*}. For a more complete error analysis, we refer to Appendix \ref{app:error}, and for samples of the model, to Appendix \ref{app:samples}. In Appendix \ref{app:results}, we break down the results for each of the functions in Synthesis-Bench, include the average length of the outputs of the different models, and report the most frequent syntactical errors and error intersections between statistics between the different models.

\paragraph{Scaling}
There is no compelling reason to believe that neural networks would not scale with data, model size, and compute in a similar way to other domains \citep{DBLP:journals/corr/abs-2001-08361}. Indeed, the models generally perform better with  more data, compute, and parameters, even though the largest model we trained was not the best. 
This may be due to insufficient  training data  or a sub-optimal training procedure e.g. insufficient  updates.
However, unlike other domains, code  quality is evaluated in a binary fashion, correct or illegal which  may cause sharp accuracy curves.

\paragraph{Limits}
Our best model can correctly compile less than half of the examples in the IO evaluation. It is, therefore,  far from being usable in practice. Furthermore, we have no control over the output space, and we operate on small functions instead of entire programs. In this work, apart from using code tokenizers and IO evaluation, we have not included any domain knowledge. Given the large amount of prior syntatctic and semantic information available for source and target, an obvious next step is to 
incorporate this in to the translation scheme.

\paragraph{Ethical concerns regarding crawling data} Finally, we remark on  the potential ethical and legal implications of training models on Github data, an emerging topic in the machine learning for code community due to the release of OpenAI's Codex \cite{DBLP:journals/corr/abs-2107-03374}. This is not an immediate concern, as we  limit our training  to an already published dataset. Any future work, however,  which  accesses 
Github code, needs to address these issues  and not to access  repositories with restrictive licenses.

\section{Conclusion}
\label{sec:con}
We conclude that our neural compilation approach shows that sequence-to-sequence deep learning models can, indeed, learn to compile end-to-end. Nevertheless, the performance is far from ideal and the restrictions make it still far from being usable in practice. The task presents many challenges, such as output length or hard syntactic and correctness constraints, that were not explicitly tackled in this work.

As future work, we suggest \begin{enumerate*}
\item scaling up our approach, in terms of data, compute, and model parameters, \item investigating how to incorporate domain knowledge in form of inductive biases or alternative data representations and inputs, and \item researching unsupervised techniques to leverage unlabelled (i.e., not parallel) code or assembler \end{enumerate*}.

\newpage



\medskip
\bibliography{references}

\newpage
\appendix



\section{Preprocessing pipeline}
\label{app:preprocessing}
Our preprocessing pipeline is composed of the following steps:
\begin{enumerate}
    \item Compilation: We use the GCC compiler to compile the C code into x86 assembler. We do not  apply any optimizations (-O0).
    \item Boilerplate removal: 
    We remove the headers and type and constant definitions. Likewise, we remove the header and footer of the assembler. In both cases, we believe those inject noise and make sequences longer than need be.
    \item Pre-tokenization: We use the GCC C and x86 assembler (GAS) tokenizers with the Pygments\footnote{\url{https://pygments.org/}} library. In C, new lines are meaningless and just used to make code more human readable, but in GAS end of lines delimits the end of each instruction. Thus, in the latter we replace end of lines by a special token \texttt{<newline>}.
    \item Length filtering: Due to computational restrictions and potentially easing the task, we discard the (C, assembler) pairs such that when summing the length of tokens of the C code and assembler we get more than 314 tokens. 
    \item Train-valid-test split: We randomly split the pairs into training, validation, and test sets, with 2k programs for validation and test and the rest for training.
    \item Subword tokenization: We use subword encoding to automatically split tokens into further tokens based on n-gram frequencies in the train set. Specifically, \texttt{subword-nmt} \citep{DBLP:journals/corr/SennrichHB15} .\footnote{\url{https://github.com/rsennrich/subword-nmt}} This has the benefit of decreasing the vocabulary size while making out-of-vocabulary tokens virtually impossible (since unknown tokens can be reconstructed from ASCII characters or other subwords present in the vocabulary).
    \item Formatting: We write each C and assembler programs in plain text files, such that we have one program for each line.
\end{enumerate}

\section{Data}
\label{app:data}

\begin{table}[hbt!]
\centering
\begin{tabular}{lrrrr}
\hline
\textsc{Dataset} & \textsc{\# Programs} &  Filter & \textsc{\# Kept Programs} & \textsc{IO} \\
\hline
Angha-Par500k & 1.044M & Max. length & 500k  & \xmark \\
Angha-Par250k & 1.044M & Max. length + random & 250k &  \xmark\\
Synthesis-Bench & 112 & Manual (difficulty) &  64 & \cmark \\

\end{tabular}
\caption{\label{tab:corpus_datasets} Used datasets, original number of programs, filter criteria, number of kept programs after filtering, and whether they have input/output examples (which only Synthesis-Bench does). The AnghaPar corpus was filtered with a maximum combined (C + assembler) length of 314 tokens. The 250k subset was further subsampled randomly. Finally, the Synthesis-Bench was built from a manual selection of 64 functions from the original benchmark, based on implementation difficulty.}
\end{table}

\begin{table}[hbt!]
\centering
\begin{tabular}{lrrrr}
\hline \textsc{Split} & \textsc{Programs} & \textsc{Tokens C (avg)} & \textsc{Tokens ASM (avg)} \\ 
\hline
Angha-Par500k  Train	& 500,439	& 22,653,480 (45.27)	& 65,910,582 (131.71)\\
Angha-Par250k  Train	& 250,000	& 11,281,616 (45.12)	& 32,992,914 (131.97)\\
Angha-Par  Valid	& 1,000	& 45,737 (45.74)	& 132,424 (132.42) \\
Angha-Par  Test	& 1,000	& 44,643 (44.64)	& 132,446 (132.37)\\

\end{tabular}
\caption{\label{tab:dataset-splits} Dataset splits. assembler code has almost 3x tokens than its corresponding C code.}
\end{table}

\begin{table}[H]
\centering
\scalebox{1.0}{
\begin{tabular}{lccc}
\hline
\textsc{Vocab} & \textsc{Subw/tok C (avg len)} & \textsc{Subw/token ASM (avg len)} & \textsc{Coverage} \\
\hline
4k & 1.55 (69.85) & 1.14 (149.85) & 100\% \\
8k & 1.42 (64.22) & 1.10 (144.99) & 100\% \\
16k & 1.33 (59.96) & 1.08 (143.12) & 100\% \\

\end{tabular}} 
\caption{\label{tab:subwords}
Subwords per token. All vocabularies have a coverage of 100\% (i.e., no unknowns) since they include all ASCII characters. C code length is more sensitive to the vocabulary size, since it has a larger vocabulary (e.g., identifiers, except procedure names, are translated as memory positions or registers). There is a clear trade-off between sequence length and vocabulary size.
}
\end{table}

\section{Experiments}
\label{app:experiments}

We experiment with the following models:
\begin{itemize}
    \item Transformer-Small: The \textit{small} model follows the \texttt{transformer\_iwslt\_de\_en} configuration in Fairseq, that is, 6 encoder layers and 6 decoder layers, an embedding size of 512 and 4 attention heads.
      \item Transformer-Big (base): The \textit{big} model follows the \texttt{transformer\_wmt\_en\_de\_big\_t2t} configuration in Fairseq, with 6 encoder layers and 6 decoder layers, an embedding size of 1024 and 16 attention heads
    \begin{itemize}
        \item -50\% data: Transformer-Big trained with Angha-Par250k instead of Angha-Par500k.
        \item -1/2x vocab: Transformer-Big trained with a vocabulary of 4k tokens (instead of 8k tokens).
        \item +1/2x vocab: Transformer-Big trained with a vocabulary of 16k tokens (instead of 8k tokens).
        \item +1e2x weight-decay: Transformer-Big further regularized (a weight decay of 0.01 instead of  0.0001)
        \item +1/2 epochs: Transformer-Big trained for a total of 10 epochs (instead of 5).
    \end{itemize}
    \item Transformer-Med: The medium-size model roughly follows the Transformer-Big configuration, but with 8 attention heads (instead of 16) and a Feed-Forward hidden size of 2048 (instead of 4096).
    \item Transformer-Big+: This model has the same configuration as Transformer-Big, but with 2 additional layers for both the encoder and the decoder.
\end{itemize}

\section{Expanded results}
\label{app:results}

We report the fine-grained IO evaluation for the best model in Table \ref{tab:fine_grained}, together with other metrics to ease the analysis of the results. Specifically, apart from the aforementioned syntactic accuracy and BLEU scores, we also report: \begin{enumerate*}

    \item \textit{LOC}: Lines of Code, the number of lines of the C implementation.
    \item Tokens: The number of tokens of the C implementation.
    \item \textit{Cyclo}: The \textit{cyclomatic complexity}: 
    $\texttt{Cyclomatic complexity} = E - N + 2\times P$
     where $E$ is the number of edges in the flow graph, $N$ is the number of nodes in the flow graph, and $P$ is the number of nodes that have exit points. 
    \item Params: The number of parameters of the C function.
    \item Pointers: The number of pointer parameters (typically arrays) of the C function.
\end{enumerate*}

Finally, Tables \ref{tab:corr}, \ref{tab:lengths}, \ref{tab:syntax_errors}, \ref{tab:io_errors} show the correlations between IO errors and other metrics, the mean output length of each model, the most frequent syntactical errors, and the most frequent IO errors, respectively. 
\newpage


\small
\centering
\begin{longtable}{lrrrrrrrr} 

\hline
\textsc{Func} & \textsc{IO} & \textsc{Syntax} & \textsc{BLEU} &  \textsc{LOC} & \textsc{TOKENS} & \textsc{CYCLO} & \textsc{PARAMS} & \textsc{POIN} \\
\hline
\texttt{add} &  \xmark &  \cmark &  85.23 &  6 &  39 &  2 &  3 &  1\\
\texttt{array\_inc} &  \xmark &  \cmark &  87.6 &  5 &  34 &  2 &  2 &  1\\
\texttt{array\_prod} &  \cmark &  \cmark &  97.96 &  7 &  42 &  2 &  2 &  1\\
\texttt{array\_sum} &  \cmark &  \cmark &  97.8 &  7 &  42 &  2 &  2 &  1\\
\texttt{binary\_digits} &  \cmark &  \cmark &  97.27 &  8 &  31 &  2 &  1 &  0\\
\texttt{binary\_mul\_sum} &  \xmark &  \xmark &  50.59 &  8 &  66 &  2 &  3 &  2\\
\texttt{clamp} &  \xmark &  \cmark &  96.77 &  7 &  45 &  3 &  2 &  1\\
\texttt{collatz} &  \cmark &  \cmark &  98.2 &  12 &  54 &  3 &  1 &  0\\
\texttt{count\_odds} &  \cmark &  \cmark &  87.58 &  9 &  52 &  3 &  2 &  1\\
\texttt{cube\_in\_place} &  \xmark &  \xmark &  65.55 &  5 &  47 &  2 &  2 &  1\\
\texttt{digit\_prod} &  \xmark &  \xmark &  59.63 &  9 &  38 &  2 &  1 &  0\\
\texttt{digits} &  \cmark &  \cmark &  82.32 &  8 &  31 &  2 &  1 &  0\\
\texttt{diveq} &  \xmark &  \cmark &  75.31 &  5 &  41 &  2 &  3 &  2\\
\texttt{diveq\_sca} &  \xmark &  \cmark &  82.79 &  5 &  37 &  2 &  3 &  1\\
\texttt{dot} &  \xmark &  \cmark &  97.5 &  7 &  51 &  2 &  3 &  2\\
\texttt{elementwise\_} &  \xmark &  \xmark &  3.25 &  15 &  122 &  4 &  4 &  3\\
\hspace{1mm} \texttt{\_sum\_of\_} &  &  &  &  &  &   &   &  \\
\hspace{1mm} \texttt{\_negated\_sum\_} &  &  &  &  &  &   &   &  \\
\hspace{1mm} \texttt{\_and\_max} &  &  &  &  &  &   &   &  \\
\texttt{eq} &  \xmark &  \xmark &  81.47 &  9 &  57 &  3 &  3 &  2\\
\texttt{fact} &  \cmark &  \cmark &  96.94 &  8 &  31 &  2 &  1 &  0\\
\texttt{fact\_fact} &  \cmark &  \cmark &  96.94 &  8 &  31 &  2 &  1 &  0\\
\texttt{fib\_n} &  \cmark &  \cmark &  97.42 &  10 &  46 &  2 &  1 &  0\\
\texttt{fourth\_in\_place} &  \xmark &  \xmark &  45.37 &  6 &  57 &  2 &  2 &  1\\
\texttt{int\_sqrt} &  \cmark &  \cmark &  86.34 &  9 &  43 &  2 &  1 &  0\\
\texttt{last\_elem} &  \cmark &  \cmark &  97.8 &  7 &  42 &  2 &  2 &  1\\
\texttt{last\_zero\_idx} &  \cmark &  \cmark &  98.04 &  9 &  50 &  3 &  2 &  1\\
\texttt{length} &  \xmark &  \xmark &  41.35 &  1 &  14 &  1 &  2 &  1\\
\texttt{max} &  \xmark &  \xmark &  79.59 &  11 &  63 &  3 &  2 &  1\\
\texttt{max\_elt} &  \xmark &  \xmark &  87.36 &  9 &  53 &  3 &  2 &  1\\
\texttt{min} &  \xmark &  \xmark &  80.04 &  11 &  63 &  3 &  2 &  1\\
\texttt{min\_elt} &  \xmark &  \xmark &  88.04 &  9 &  53 &  3 &  2 &  1\\
\texttt{min\_so\_far\_} &  \xmark &  \cmark &  0.0 &  18 &  157 &  6 &  4 &  3\\
\hspace{1mm} \texttt{\_subtracted} &  &  &  &  &  &   &   &  \\
\texttt{mirror\_image} &  \xmark &  \xmark &  77.16 &  9 &  61 &  3 &  3 &  2\\
\texttt{muleq} &  \xmark &  \cmark &  73.78 &  5 &  41 &  2 &  3 &  2\\
\texttt{muleq\_sca} &  \xmark &  \cmark &  85.0 &  5 &  37 &  2 &  3 &  1\\
\texttt{negate} &  \xmark &  \cmark &  87.71 &  5 &  38 &  2 &  2 &  1\\
\texttt{pluseq} &  \xmark &  \xmark &  76.4 &  5 &  41 &  2 &  3 &  2\\
\texttt{prod\_elts} &  \cmark &  \cmark &  97.96 &  7 &  42 &  2 &  2 &  1\\
\texttt{prod\_n\_squared} &  \cmark &  \cmark &  97.66 &  8 &  39 &  2 &  1 &  0\\
\texttt{prod\_sq\_elts} &  \xmark &  \xmark &  85.46 &  8 &  49 &  2 &  2 &  1\\
\texttt{replace\_first} &  \xmark &  \xmark &  79.77 &  9 &  62 &  3 &  2 &  1\\
\texttt{replace\_last} &  \xmark &  \xmark &  79.89 &  9 &  62 &  3 &  2 &  1\\
\texttt{reverse} &  \xmark &  \xmark &  55.01 &  7 &  62 &  2 &  2 &  1\\
\texttt{reverse\_int} &  \xmark &  \xmark &  61.04 &  9 &  37 &  2 &  1 &  0\\
\texttt{search} &  \cmark &  \cmark &  95.23 &  9 &  59 &  4 &  3 &  1\\
\texttt{sort} &  \xmark &  \xmark &  33.63 &  9 &  84 &  4 &  2 &  1\\
\texttt{subeq} &  \xmark &  \xmark &  74.26 &  5 &  41 &  2 &  3 &  2\\
\texttt{subeq\_sca} &  \xmark &  \cmark &  89.73 &  5 &  37 &  2 &  3 &  1\\
\texttt{subtract\_of\_} &  \xmark &  \xmark &  46.71 &  8 &  82 &  3 &  4 &  3\\
\hspace{1mm} \texttt{\_min\_reverse} &  &  &  &  &  &   &   &  \\
\texttt{sum\_abs} &  \xmark &  \xmark &  59.81 &  7 &  57 &  3 &  2 &  1\\
\texttt{sum\_elts} &  \cmark &  \cmark &  97.8 &  7 &  42 &  2 &  2 &  1\\
\texttt{sum\_n} &  \cmark &  \cmark &  96.74 &  8 &  30 &  2 &  1 &  0\\
\texttt{sum\_n\_squared} &  \cmark &  \cmark &  92.65 &  8 &  32 &  2 &  1 &  0\\
\texttt{sum\_of\_lists\_} &  \xmark &  \xmark &  18.44 &  13 &  105 &  4 &  3 &  2\\
\hspace{1mm} \texttt{\_multiplied\_} &  &  &  &  &  &   &   &  \\
\hspace{1mm} \texttt{\_after\_dividing\_} & &   &  &  &   &  &  &  \\
\hspace{1mm} \texttt{\_by\_three} & &   &  &  &   &  &  &  \\
\texttt{sum\_of\_positives} &  \xmark &  \xmark &  44.02 &  10 &  91 &  4 &  4 &  3\\
\texttt{sum\_of\_squares} &  \cmark &  \cmark &  98.27 &  7 &  47 &  2 &  2 &  1\\
\texttt{triangle\_prod} &  \cmark &  \cmark &  97.95 &  9 &  51 &  3 &  1 &  0\\
\texttt{triangle\_sum} &  \cmark &  \cmark &  97.86 &  9 &  51 &  3 &  1 &  0\\
\texttt{vadd} &  \xmark &  \xmark &  73.97 &  5 &  50 &  2 &  4 &  3\\
\texttt{vcopy} &  \xmark &  \cmark &  84.61 &  5 &  41 &  2 &  3 &  2\\
\texttt{vfill} &  \xmark &  \cmark &  96.03 &  5 &  37 &  2 &  3 &  1\\
\texttt{vmul} &  \xmark &  \xmark &  72.34 &  5 &  50 &  2 &  4 &  3\\
\texttt{vneg} &  \xmark &  \cmark &  87.71 &  5 &  38 &  2 &  2 &  1\\
\texttt{voffset} &  \xmark &  \cmark &  85.23 &  5 &  37 &  2 &  3 &  1\\
\texttt{vscal~} &  \xmark &  \cmark &  85.0 &  5 &  37 &  2 &  3 &  1\\
\texttt{vsub} &  \xmark &  \xmark &  71.26 &  5 &  50 &  2 &  4 &  3\\
\caption{\label{tab:fine_grained}
Best model in Synthesis-Bench: IO and syntactic accuracy and BLEU of the model output, and cyclomatic complexity, n. of parameters and pointer parameters of the C function. 
}
\end{longtable}


\normalsize

\section{Expanded error analysis}
\label{app:error}

\begin{flushleft}
Focusing on the outputs of the best model, we observe:
\end{flushleft}

\begin{itemize}
    \item When the model has one correct IO test in a given function, it is likely that the others will be also correct, as shown in Table \ref{tab:io_errors}. The probability of generating a program that only passes one unit test by chance is, indeed, very low.
    \item After manually inspecting the most frequent syntactical errors (Table \ref{tab:syntax_errors}), we find that most of these occur because the output finishes prematurely. For instance, it is common to find outputs with operators with unbalanced parentheses as the last instruction, not because the model has not learned the syntax, but because the decoding terminated in the middle of the program. This occurs when outputs are long and the model  predicts the end of the program special token prematurely.
    \item In our experiments, IO accuracy does not correlate with cyclomatic complexity, as shown in Table \ref{tab:corr}. We see two potential reasons for that, namely, \begin{enumerate*}
        \item in Synthesis-Bench there are not enough functions to observe sufficient variability in cyclomatic complexity to observe the expected correlation, 
        or \item the sources of the errors are more simple (e.g., the mere presence of an array) than the complexity captured by cyclomatic complexity.
    \end{enumerate*}. In fact, the number of function parameters and the number of points seems to be indeed negatively correlated with the IO accuracy. Thus, we conclude that the more function parameters and more pointers, the more difficult is for neural models to correctly interpret C and generate the corresponding assembler. Finally, with no surprise, syntactical accuracy and BLEU score positively correlate with IO accuracy, since correct solutions are clearly syntactically correct and, with a lesser degree, lexically similar to the GCC solution. However, the correlation is not strong enough for these metrics to be used as reliable proxies of the IO accuracy in case unit tests are not available.
    \item All models fail in the same functions: Table \ref{tab:intersections} shows that the intersection of IO errors between the different models is almost full, meaning that errors are related to some intrinsic difficulty of these functions (at least to neural compilers) and not to randomness in the training process. 
    \item Model outputs do appear like GCC outputs, but with some artifacts such as unnecessary nop operations in some cases (see supplementary material).
    \item We observe some trivial errors. For instance, \texttt{true} and \texttt{false} (boolean values from \texttt{stdbool}) are confused with  variable names. If they are manually replaced with 1 and 0, the models usually generate a correct output.
\end{itemize}

\begin{table}[hbt!]
\centering
\begin{tabular}{lc}
\hline
\textsc{Metric} & \textsc{Correlation (p-value)} \\
\hline

Syntax  &    \textbf{0.597 (1.92E-07)} \\
BLEU     &     \textbf{0.536 (4.96E-06)} \\
LOC    &    0.174 (1.69E-01) \\
Tokens     & \textbf{-0.269 (3.13E-02)} \\
Cyclo     & -0.106 (4.04E-01) \\
Params    & \textbf{-0.607 (1.04E-07)} \\
Pointers &    \textbf{-0.573 (7.56E-07)} \\
\end{tabular}%
\caption{\label{tab:corr}
Pearson correlations between different metrics (syntactical accuracy, BLEU score, lines of code and number of tokens in the C implementation, cyclomatic complexity of the C implementation, number of parameters in the C function, and number of pointer parameters in the C function) and IO accuracy. Bold values are statistically significant.
}
\end{table}

\begin{table}
\centering
\scalebox{1.0}{
\begin{tabular}{lr}
\hline
\textsc{Model} & \textsc{Avg Output Length} \\
\hline
Transformer-Small & \textbf{162.29} \\
Transformer-Med & 124.94 \\
Transformer-Big  & 124.61 \\
\hspace{3mm} - 50\% data & 125.00 \\
\hspace{3mm} - 1/2x vocab & 127.22 \\
\hspace{3mm} + 1/2x vocab  & 124.13 \\
\hspace{3mm} + 1e2x weight-decay & 124.74 \\
\hspace{3mm} + 1/2x epochs &124.59 \\
Transformer-Big+ & 124.99 \\ 

\hline
Ground truth & 132.37 \\
\end{tabular}} 
\caption{\label{tab:lengths}
Average length of the output of the different models in the Angha-Par test, vs. the ground truth (GCC) one. 
}
\end{table}

\begin{table*}
\centering
\scalebox{1.0}{
\begin{tabular}{l}
\hline
\textsc{Error}  \\
\hline
     \texttt{open CFI at the end of file; missing .cfi\_endproc directive} \\ 
      \texttt{expecting operand after ','; got nothing} \\
      \texttt{unbalanced brackets in operand 1.} \\
      \texttt{number of operands mismatch for `mov'} \\
      \texttt{number of operands mismatch for `add'} \\
      \texttt{unbalanced brackets in operand 2.} \\
      \texttt{bad or irreducible absolute expression} \\
      \texttt{CFI instruction used without previous .cfi\_startproc} \\
     \texttt{junk at end of line, first unrecognised character is `\%'} \\
      \texttt{symbol `.L3' is already defined} \\
       \texttt{number of operands mismatch for `cmp'} \\
       \texttt{symbol `.L5' is already defined} \\ 
       \texttt{number of operands mismatch for `movq'} \\
       \texttt{number of operands mismatch for `lea'} \\
      \texttt{.cfi\_endproc without corresponding .cfi\_startproc} \\
       \texttt{symbol `.L4' is already defined} \\
       \texttt{operand type mismatch for `sar'} \\
       \texttt{number of operands mismatch for `pop'} \\
       \texttt{number of operands mismatch for `sal'} \\
       \texttt{number of operands mismatch for `pxor'} \\
       \texttt{number of operands mismatch for `movslq'} \\
       \texttt{.size expression for sum\_n does not evaluate to a constant} \\

\end{tabular}} 
\caption{\label{tab:syntax_errors}
Frequent syntactical errors (sorted by frequency).
}
\end{table*}

\begin{table*}
\centering
\scalebox{1.0}{
\begin{tabular}{lr}
\hline
\textsc{Error}  \\
\hline
Syntax error & 27 \\
Compiled but 0 tests passed & 15 \\
Compiled but only 1 test passed & 1 \\
Compiled but more than 1 test passed & 0 
\end{tabular}} 
\caption{\label{tab:io_errors}
IO error typology for the best model.
}
\end{table*}

\begin{table*}
\centering
\scalebox{1.0}{
\begin{tabular}{lr}
\hline
\textsc{Model} & \textsc{Intersections} \\
\hline
Transformer-Small & 0/0 \\
Transformer-Med & 18/19 \\
Transformer-Big  & 19/19 \\
\hspace{3mm} - 50\% data & 13/13 \\
\hspace{3mm} - 1/2x vocab & 19/19 \\
\hspace{3mm} + 1/2x vocab  & 20/20 \\
\hspace{3mm} + 1e2x weight-decay & 18/18 \\
\hspace{3mm} + 1/2x epochs & 21/21 \\
Transformer-Big+ & 19/19 \\ 

\end{tabular}} 
\caption{\label{tab:intersections}
Intersections between the correct outputs of each model and the correct outputs of the best one (Transformer-Big + 1/2x epochs). 
}
\end{table*}

\clearpage

\section{Output samples}
\label{app:samples}

\begin{flushleft}

We include some model output samples, together with the reference GCC output and the original C, selected as follows: 
\end{flushleft}
\begin{itemize}

    \item All outputs belong to the best model, specifically they are one of the top 5 hypothesis.
    \item All outputs belong to Synthesis-Bench.
    \item They have not been randomly sampled, or cherry-picked to show the best outputs only, but manually curated to try to provide representative outputs of different cases, which may not appear proportionally.
\end{itemize}





\begin{figure}[H]
\noindent\begin{minipage}[t]{.3\textwidth}
\begin{lstlisting}[caption=C input,frame=tlrb, basicstyle=\tiny]{Name}
int triangle_sum(int n) {
  int r = 0;
  for (int i = 1; i < n;
       ++i) {
    for (int m = 1; m < i;
         ++m) {
      r += m;
    }
  }
  return r;
}

\end{lstlisting}
\end{minipage}\hfill
\begin{minipage}[t]{.3\textwidth}
\begin{lstlisting}[caption=Model assembler,frame=tlrb, basicstyle=\tiny]{Name}
triangle_sum: 
 .LFB0: 
 .cfi_startproc 
 pushq %rbp 
 .cfi_def_cfa_offset 16 
 .cfi_offset 6, -16 
 movq %rsp, %rbp 
 .cfi_def_cfa_register 6 
 movl %edi, -20(%rbp) 
 movl $0, -12(%rbp) 
 movl $1, -8(%rbp) 
 jmp .L2 
 .L5: 
 movl $1, -4(%rbp) 
 jmp .L3 
 .L4: 
 movl -4 (%rbp), %eax 
 addl %eax, -12(%rbp) 
 addl $1, -4(%rbp) 
 .L3: 
 movl -4(%rbp), %eax 
 cmpl -8(%rbp), %eax 
 jl .L4 
 addl $1, -8(%rbp) 
 .L2: 
 movl -8(%rbp), %eax 
 cmpl -20(%rbp), %eax 
 jl .L5 
 movl -12(%rbp), %eax 
 popq %rbp 
 .cfi_def_cfa 7, 8 
 ret 
 .cfi_endproc
\end{lstlisting}
\end{minipage}\hfill
\begin{minipage}[t]{.3\textwidth}
\begin{lstlisting}[caption=GCC assembler,frame=tlrb, basicstyle=\tiny]{Name}
triangle_sum:
.LFB0:
 .cfi_startproc
 endbr64
 pushq %rbp
 .cfi_def_cfa_offset 16
 .cfi_offset 6, -16
 movq %rsp, %rbp
 .cfi_def_cfa_register 6
 movl %edi, -20(%rbp)
 movl $0, -12(%rbp)
 movl $1, -8(%rbp)
 jmp  .L2
.L5:
 movl $1, -4(%rbp)
 jmp  .L3
.L4:
 movl -4(%rbp), %eax
 addl %eax, -12(%rbp)
 addl $1, -4(%rbp)
.L3:
 movl -4(%rbp), %eax
 cmpl -8(%rbp), %eax
 jl   .L4
 addl $1, -8(%rbp)
.L2:
 movl -8(%rbp), %eax
 cmpl -20(%rbp), %eax
 jl   .L5
 movl -12(%rbp), %eax
 popq %rbp
 .cfi_def_cfa 7, 8
 ret
 .cfi_endproc
\end{lstlisting}
\end{minipage}
    \caption{Correct output sample in which the system output is almost identical to the GCC one. The only difference is the inclusion of the new instruction \texttt{endbr64}, which is used for security reasons and is executed as a \texttt{NOP}s in targets that do not support it.
    } 
    \label{fig:sample1}
\end{figure}
\clearpage


\begin{figure}[H]
\noindent\begin{minipage}[t]{.3\textwidth}
\begin{lstlisting}[caption=C input,frame=tlrb, basicstyle=\tiny]
int fib_n(int n) {
  int i = 1;
  int r = 1;
  while (n > 1) {
    i = r - i;
    r = i + r;
    n = n - 1;
  }
  return r;
}

\end{lstlisting}
\end{minipage}\hfill
\begin{minipage}[t]{.3\textwidth}
\begin{lstlisting}[caption=Model assembler,frame=tlrb, basicstyle=\tiny]
fib_n: 
 .LFB0: 
 .cfi_startproc 
 pushq %rbp 
 .cfi_def_cfa_offset 16 
 .cfi_offset 6 , -16 
 movq %rsp , %rbp 
 .cfi_def_cfa_register 6 
 movl %edi , -20 ( %rbp ) 
 movl $1 , -8 ( %rbp ) 
 movl $1 , -4 ( %rbp ) 
 jmp .L2 
 .L3: 
 movl -4 ( %rbp ), %eax 
 subl -8 ( %rbp ), %eax 
 movl %eax , -8 ( %rbp ) 
 movl -8 ( %rbp ), %edx 
 movl -4 ( %rbp ), %eax 
 addl %edx , %eax 
 movl %eax , -4 ( %rbp ) 
 subl $1 , -20 ( %rbp ) 
 .L2: 
 cmpl $1 , -20 ( %rbp ) 
 jg .L3 
 movl -4 ( %rbp ), %eax 
 popq %rbp 
 .cfi_def_cfa 7 , 8 
 ret 
 .cfi_endproc 


\end{lstlisting}
\end{minipage}\hfill
\begin{minipage}[t]{.3\textwidth}
\begin{lstlisting}[caption=GCC assembler,frame=tlrb, basicstyle=\tiny]
fib_n:
.LFB0:
 .cfi_startproc
 endbr64
 pushq %rbp
 .cfi_def_cfa_offset 16
 .cfi_offset 6, -16
 movq %rsp, %rbp
 .cfi_def_cfa_register 6
 movl %edi, -20(%rbp)
 movl $1, -8(%rbp)
 movl $1, -4(%rbp)
 jmp  .L2
.L3:
 movl -4(%rbp), %eax
 subl -8(%rbp), %eax
 movl %eax, -8(%rbp)
 movl -8(%rbp), %eax
 addl %eax, -4(%rbp)
 subl $1, -20(%rbp)
.L2:
 cmpl $1, -20(%rbp)
 jg   .L3
 movl -4(%rbp), %eax
 popq %rbp
 .cfi_def_cfa 7, 8
 ret
 .cfi_endproc
\end{lstlisting}
\end{minipage}
    \caption{Correct compilation of an iterative Fibonacci implementation, in which the system output subtly differs from the GCC one.}
    \label{fig:sample2}
\end{figure}

\begin{figure}[H]

\noindent\begin{minipage}[t]{.3\textwidth}
\begin{lstlisting}[caption=C input,frame=tlrb, basicstyle=\tiny]
int search(int *arr,
           int v,
           int n) {
  int idx = -1;
  for (int i = 0; i < n;
       ++i) {
    if (idx == -1 &&
        arr[i] == v) {
      idx = i;
    }
  }
  return idx;
}

\end{lstlisting}
\end{minipage}\hfill
\begin{minipage}[t]{.3\textwidth}
\begin{lstlisting}[caption=Model assembler,frame=tlrb,   basicstyle=\tiny,]
search: 
 .LFB0: 
 .cfi_startproc 
 pushq %rbp 
 .cfi_def_cfa_offset 16 
 .cfi_offset 6, -16 
 movq %rsp, %rbp 
 .cfi_def_cfa_register 6 
 movq %rdi, -24(%rbp) 
 movl %esi, -28(%rbp) 
 movl %edx , -32(%rbp) 
 movl $-1, -8(%rbp) 
 movl $0, -4(%rbp) 
 jmp .L2 
 .L4: 
 cmpl $-1, -8(%rbp) 
 jne .L3 
 movl -4(%rbp), %eax 
 movslq %eax, %rdx 
 movq -24(%rbp), %rax 
 addq %rdx, %rax 
 movl (%rax), %eax 
 cmpl -28(%rbp), %eax 
 jne .L3 
 movl -4(%rbp), %eax 
 movl %eax, -8(%rbp) 
 .L3: 
 addl $1, -4(%rbp) 
 .L2: 
 movl -4(%rbp), %eax 
 cmpl -32(%rbp), %eax 
 jl .L4 
 movl -8(%rbp), %eax 
 popq %rbp 
 .cfi_def_cfa 7, 8 
 ret 
 .cfi_endproc
 

\end{lstlisting}
\end{minipage}\hfill
\begin{minipage}[t]{.3\textwidth}
\begin{lstlisting}[caption=GCC assembler,frame=tlrb, basicstyle=\tiny]
search:
.LFB0:
 .cfi_startproc
 endbr64
 pushq %rbp
 .cfi_def_cfa_offset 16
 .cfi_offset 6, -16
 movq %rsp, %rbp
 .cfi_def_cfa_register 6
 movq %rdi, -24(%rbp)
 movl %esi, -28(%rbp)
 movl %edx, -32(%rbp)
 movl $-1, -8(%rbp)
 movl $0, -4(%rbp)
 jmp  .L2
.L4:
 cmpl $-1, -8(%rbp)
 jne  .L3
 movl -4(%rbp), %eax
 cltq
 leaq 0(,%rax,4), %rdx
 movq -24(%rbp), %rax
 addq %rdx, %rax
 movl (%rax), %eax
 cmpl %eax, -28(%rbp)
 jne  .L3
 movl -4(%rbp), %eax
 movl %eax, -8(%rbp)
.L3:
 addl $1, -4(%rbp)
.L2:
 movl -4(%rbp), %eax
 cmpl -32(%rbp), %eax
 jl   .L4
 movl -8(%rbp), %eax
 popq %rbp
 .cfi_def_cfa 7, 8
 ret
 .cfi_endproc

\end{lstlisting}
\end{minipage}
    \caption{Incorrect output sample (top 3 hypothesis of the best model in the  \texttt{search} function) that passes only some (5/9) of the IO examples. This is relatively uncommon, since usually either it passes all tests or none, and when one hypothesis in the top 5 ones obtains better than half of the examples right, it is likely that some other hypothesis will pass all tests, as happened in this case (top 1 and 2 hypothesis were correct).}
    \label{fig:sample3}
\end{figure}

\begin{figure}[H]
\noindent\begin{minipage}[t]{.3\textwidth}
\begin{lstlisting}[caption=C input,frame=tlrb, basicstyle=\tiny]
void min_so_far_subtracted
(int *arr, int *arr2,
int *result, int n) {
  int min_so_far_array[n];
  int min_so_far = 1;

  for (int i = 0; i < n;
  i ++) {
    min_so_far =
     (min_so_far < arr[i])
     ? min_so_far : 1;
    min_so_far_array[i] =
     min_so_far;
  }
  int max_of_min_so_far
    _and_other_array[n];
  for (int i = 0; i < n;
  i ++) {
    int m1 = min_so_far
      _array[i];
        int m2 = arr2[i];
        int max = (m1 > m2
          ? m1 : m2);
    max_of_min_so_far
      _and_other_array[i]
      = max;
  }
  for (int i = 0; i < n;
       i ++) {
          result[i] =
            min_so_far
            _array[i] -
            max_of_min_so
            _far_and_
            other_array
            [i];
  }
}

\end{lstlisting}
\end{minipage}\hfill
\begin{minipage}[t]{.3\textwidth}
\begin{lstlisting}[caption=Model assembler,frame=tlrb, basicstyle=\tiny, breaklines=true]
min_so_far_subtracted: 
 .LFB0: 
 .cfi_startproc 
 pushq %rbp 
 .cfi_def_cfa_offset 16 
 .cfi_offset 6, -16 
 movq %rsp, %rbp 
 .cfi_def_cfa_register 6 
 movq %rdi, -8(%rbp) 
 movq %rsi, -16(%rbp) 
 movq %rdx, -24(%rbp) 
 movl %ecx, -28(%rbp) 
 movl %r8d, -32(%rbp) 
 nop 
 popq %rbp 
 .cfi_def_cfa 7, 8 
 ret 
 .cfi_endproc 
\end{lstlisting}
\end{minipage}\hfill
\begin{minipage}[t]{.3\textwidth}
\begin{lstlisting}[caption=GCC assembler,frame=tlrb, basicstyle=\tiny]
min_so_far_subtracted:
.LFB0:
 .cfi_startproc
 endbr64
 pushq %rbp
 .cfi_def_cfa_offset 16
 .cfi_offset 6, -16
 movq %rsp, %rbp
 .cfi_def_cfa_register 6
 pushq %r15
 pushq %r14
 pushq %r13
 pushq %r12
 subq $112, %rsp
 .cfi_offset 15, -24
 .cfi_offset 14, -32
 .cfi_offset 13, -40
 .cfi_offset 12, -48
 movq %rdi, -120(%rbp)
 movq %rsi, -128(%rbp)
 movq %rdx, -136(%rbp)
 movl %ecx, -140(%rbp)
 movq %fs:40, %rax
 movq %rax, -40(%rbp)
 xorl %eax, %eax
 movq %rsp, %rax
 movq %rax, %rdi
 movl -140(%rbp), %eax
 movslq  %eax, %rdx
 subq $1, %rdx
 movq %rdx, -72(%rbp)
 movslq  %eax, %rdx
 movq %rdx, %r14
 movl $0, %r15d
 movslq  %eax, %rdx
 movq %rdx, %r8
 movl $0, %r9d
 cltq
 leaq 0(,%rax,4), %rdx
 movl $16, %eax
 subq $1, %rax
 addq %rdx, %rax
 movl $16, %esi
 movl $0, %edx
 divq %rsi
 imulq   $16, %rax, %rax
 movq %rax, %rdx
 andq $-4096, %rdx
 movq %rsp, %rsi
 subq %rdx, %rsi
 movq %rsi, %rdx
.L2:
 cmpq %rdx, %rsp
 # CONTINUES FOR 168
 # INSTRUCTIONS MORE...
 .cfi_endproc
\end{lstlisting}

\end{minipage}
    \caption{Syntactically correct, yet meaningless output. This case follows a common pattern we have observed: when the input C is too complex for the model, the model just outputs a short, syntactically correct assembler that semantically has almost nothing to do with the input C. The model seems to behave like an unconditional assembler language model.}
    \label{fig:sample4}
\end{figure}

\begin{figure}[H]
\noindent\begin{minipage}[t]{.3\textwidth}
\begin{lstlisting}[caption=C input,frame=tlrb,basicstyle=\tiny]]
void norm(float *vec,
         float *o,
         int n) {
  float sum = 0.0f;
  for (int i = 0; i < n;
       ++i) {
    sum += vec[i];
  }

  for (int i = 0; i < n;
       ++i) {
    o[i] /= sum;
  }
}

\end{lstlisting}
\end{minipage}\hfill
\begin{minipage}[t]{.3\textwidth}
\begin{lstlisting}[caption=Model assembler,frame=tlrb,basicstyle=\tiny]]
subeq: 
 .LFB0: 
 .cfi_startproc 
 pushq %rbp 
 .cfi_def_cfa_offset 16 
 .cfi_offset 6, -16 
 movq %rsp, %rbp 
 .cfi_def_cfa_register 6 
 movq %rdi, -24(%rbp) 
 movq %rsi, -32(%rbp) 
 movl %edx, -36(%rbp) 
 movl $0, -4(%rbp) 
 jmp .L2 
 .L3: 
 movl -4(%rbp),%eax 
 cltq 
 leaq 0(,%rax,4), %rdx 
 movq -24(%rbp), %rax 
 addq %rdx, %rax 
 movl -4(%rbp), %edx 
 movslq %edx, %rdx 
 leaq 0 (,%rdx, 4), %rcx 
 movq -24(%rbp), %rdx 
 addq %rcx, %rdx 
 movl (%rdx), %edx 
 movl %edx, %ecx 
 movl -4(%rbp), %edx 
 movslq %edx, %rdx 
 leaq 0 (, %rdx , 4 ), %rsi 
 movq -32 (%rbp), %rdx 
 addq %rsi, %rdx 
 movl (%rdx), %edx 
 subl %edx, %ecx 
 movl %ecx, %edx 
 movl %edx, (%rax) 
 addl $1, -4(%rbp
\end{lstlisting}
\end{minipage}\hfill
\begin{minipage}[t]{.3\textwidth}
\begin{lstlisting}[caption=GCC assembler,frame=tlrb,basicstyle=\tiny]]
norm:
.LFB0:
 .cfi_startproc
 endbr64
 pushq %rbp
 .cfi_def_cfa_offset 16
 .cfi_offset 6, -16
 movq %rsp, %rbp
 .cfi_def_cfa_register 6
 movq %rdi, -24(%rbp)
 movq %rsi, -32(%rbp)
 movl %edx, -36(%rbp)
 pxor %xmm0, %xmm0
 movss   %xmm0, -12(%rbp)
 movl $0, -8(%rbp)
 jmp  .L2
.L3:
 movl -8(%rbp), %eax
 cltq
 leaq 0(,%rax,4), %rdx
 movq -24(%rbp), %rax
 addq %rdx, %rax
 movss   (%rax), %xmm0
 movss   -12(%rbp), %xmm1
 addss   %xmm1, %xmm0
 movss   %xmm0, -12(%rbp)
 addl $1, -8(%rbp)
.L2:
 movl -8(%rbp), %eax
 cmpl -36(%rbp), %eax
 jl   .L3
 movl $0, -4(%rbp)
 jmp  .L4
.L5:
 movl -4(%rbp), %eax
 cltq
 leaq 0(,%rax,4), %rdx
 movq -32(%rbp), %rax
 addq %rdx, %rax
 movss   (%rax), %xmm0
 movl -4(%rbp), %eax
 cltq
 leaq 0(,%rax,4), %rdx
 movq -32(%rbp), %rax
 addq %rdx, %rax
 divss   -12(%rbp), %xmm0
 movss   %xmm0, (%rax)
 addl $1, -4(%rbp)
.L4:
 movl -4(%rbp), %eax
 cmpl -36(%rbp), %eax
 jl   .L5
 nop
 nop
 popq %rbp
 .cfi_def_cfa 7, 8
 ret
 .cfi_endproc
\end{lstlisting}
\end{minipage}
    \caption{Syntactically incorrect output (unbalanced parentheses in the last \texttt{addl} instruction) that actually is caused by the hypothesis terminating before it should have, like most detected syntax errors.}
    \label{fig:sample5}
\end{figure}

\end{document}